%% file: root.tex
\documentclass[letterpaper, 10 pt, conference]{ieeeconf}  

\IEEEoverridecommandlockouts                              

\usepackage{graphics} 
\usepackage[caption=false]{subfig}
\usepackage{epsfig} 
\usepackage{amsmath} 
\usepackage{amssymb}  
\usepackage{censor}
\usepackage{float}
\usepackage{makecell}
\usepackage{tabularx}
\usepackage{booktabs}
\usepackage{acro}
\usepackage{placeins}
\DeclareAcronym{rl}{
  short = RL ,
  long  = Reinforcement Learning
}
\DeclareAcronym{dof}{
  short = DoF ,
  long  = Degrees of Freedom
}
\DeclareAcronym{cot}{
  short = CoT ,
  long  = Cost of Transport
}
\DeclareAcronym{sla}{
  short = SLA ,
  long  = stereolithography
}
\DeclareAcronym{pomdp}{
  short = POMDP ,
  long  = partially observable Markov decision process
}
\DeclareAcronym{ppo}{
  short = PPO ,
  long  = Proximal Policy Optimization
}
\DeclareAcronym{pd}{
  short = PD ,
  long  = Proportional-Derivative
}
\DeclareAcronym{imu}{
  short = IMU ,
  long  = Inertial Measurement Unit
}

\usepackage{tikz}
\usepackage{pgfplots}
\pgfplotsset{compat=1.18} 
\usepgfplotslibrary{groupplots, colormaps, fillbetween}
\let\labelindent\relax
\usepackage{enumitem}
\usepackage{url}

\title{\Large \bf
Reinforcement Learning-Based Control for an Inline Skating Humanoid Robot
}

\author{Ethan Marot$^{1}$, Thomas Bi$^{1}$, Clemens Schwarke$^{2}$, Victor Klemm$^{2}$, Marco Hutter$^{2}$, Raffaello D’Andrea$^{1}$
\thanks{$^{1}$Institute for Dynamic Systems and Control, ETH Zurich, Switzerland. {\tt\small \{emarot, bit, rdandrea\}@ethz.ch}}
\thanks{$^{2}$ Robotic Systems Lab, ETH Zurich, Switzerland}}

\begin{document}

\maketitle
\thispagestyle{empty}
\pagestyle{empty}

\begin{abstract}
As humanoid robots become increasingly dynamic, coupling them with reinforcement learning offers a promising approach to solving the complex, underactuated mechanics of passive inline skating. Equipping a humanoid robot with passive inline skating wheels presents an opportunity to combine the versatile agility of humanoids with the high-speed, energy-efficient locomotion strategies utilized by human skaters. In this paper, we train and deploy a reinforcement learning control policy that enables novel locomotion strategies for a humanoid robot modified to equip consumer inline skates instead of conventional feet. Unlike previous work limited to quadrupedal robots or actively driven wheels, our system allows for precise 6-DoF control of the skates to execute dynamic, edge-driven propulsion strategies. Our skating strategies emerge entirely from our reward structure, without reliance on human motion data, imitation learning, or kinematic priors. We overcome the inherent instability of passive wheels and simulation contact artifacts by utilizing different geometric wheel models (spherical and ellipsoidal) during training and validation, along with a custom success-based command curriculum and a specialized rolling reward. Consequently, our policy demonstrates up to a 50\% reduction in \ac{cot} compared to standard walking gaits. The resulting policy successfully transfers zero-shot to the physical Booster T1 hardware. Real-world deployments demonstrate dynamic balance, the ability to reject active physical perturbations, and agile locomotion strategies capable of turning at speed. A video of our results can be found at \url{https://www.youtube.com/watch?v=-_APcOS7uFo}.
\end{abstract}
\section{INTRODUCTION} \label{introduction}

Advances in humanoid hardware coupled with \ac{rl} have enabled whole-body control strategies capable of solving highly complex dynamic locomotion problems~\cite{parkourchaining, xie2025humanoid}. However, bipedal locomotion via unactuated skates remains a largely underexplored frontier that presents a complex whole-body control problem. Furthermore, while skating introduces inherent instabilities compared to standard walking or actuated legged-wheeled locomotion, it offers the potential for highly efficient locomotion across long, flat distances with minimal hardware complexity.

We focus specifically on inline skates rather than quad roller skates for our control problem. Although their co-linear wheel arrangement reduces the support polygon of each foot to a narrow line, increasing the difficulty of the task, inline skates enable higher forward velocities and are designed for long-distance traversal. Furthermore, inline skates feature no additional coupling between weight distribution and turn radius, unlike quad roller skates which utilize trucks to twist the wheels when turning. Unlike actuated wheeled-legged systems, inline skating propulsion cannot utilize direct motor torques applied at the wheel axles. Instead, the controller must execute precise 6 \ac{dof} maneuvers, strategically angling the skate edges to generate the lateral ground reaction forces necessary to propel the robot forward. This highly dynamic process is further complicated by partial observability; lacking dedicated encoders on the passive wheels, the system must infer critical states, such as wheel speed and slippage, entirely through secondary sensors to maintain dynamic balance.

\begin{figure}[t]
    \centering
    \includegraphics[width=\linewidth, trim=0 0 0 150, clip]{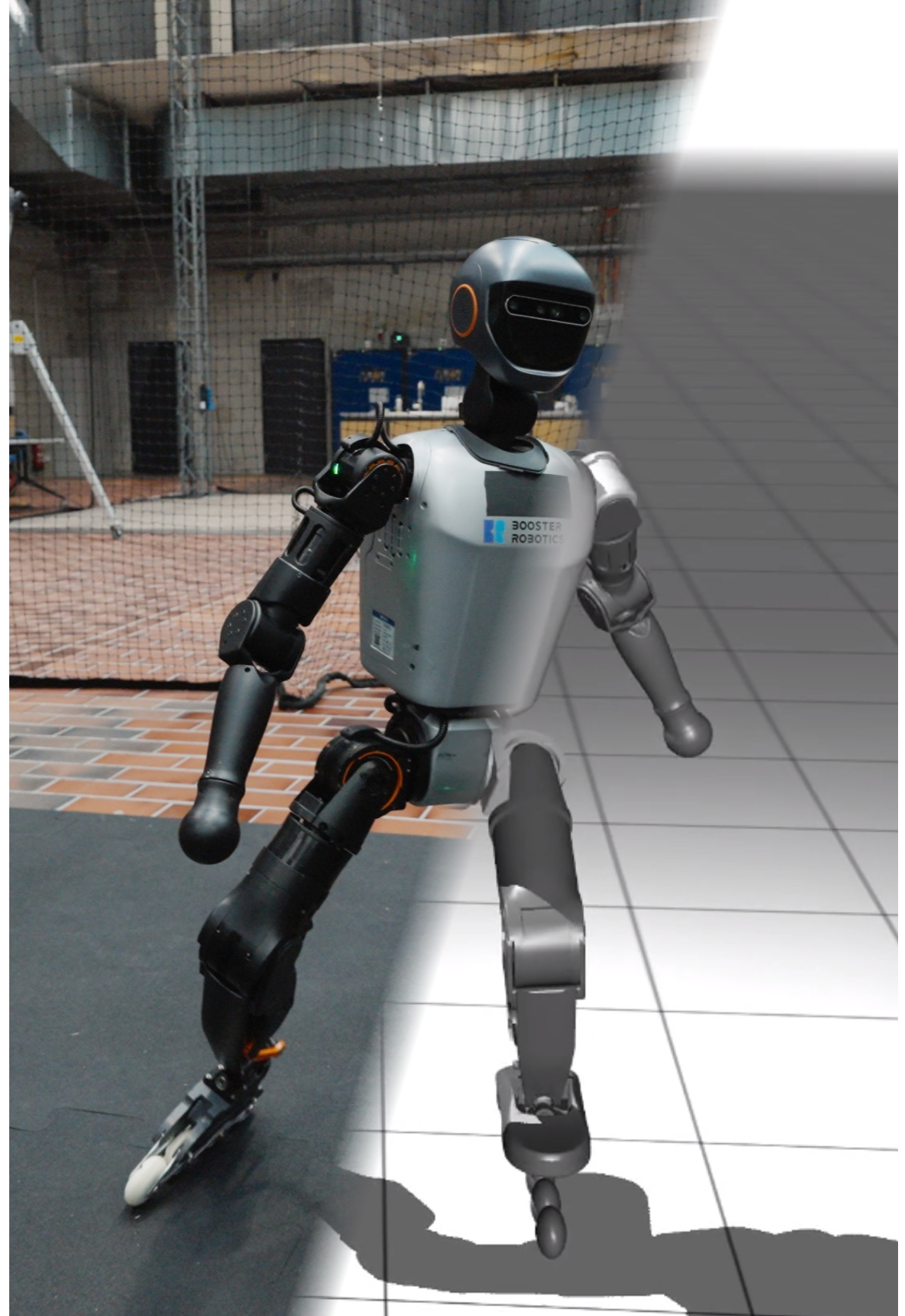}
    \caption{Zero-shot sim-to-real transfer of an inline skating policy. Shown is the MuJoCo simulation model used for validation alongside the physical Booster T1 robot. Our trained reinforcement learning controller enables the hardware to dynamically balance, glide, and turn using completely passive consumer inline skates.}
    \label{fig:showcase}
\end{figure}

Mastering these locomotion strategies could enable humanoids to participate in highly dynamic, agile domains such as inline hockey or figure skating. Additionally, further work could implement skates as modular external tools to facilitate seamless transitions between conventional stepping and high-speed rolling. Such multimodal platforms could exploit the versatility of legged locomotion to navigate rough terrain while leveraging the energy efficiency of skates to rapidly traverse flat environments.

In this work, we present, to the best of our knowledge, the first attempt to teach a humanoid robot to inline skate with emerging human-like strategies (see Fig.~\ref{fig:showcase}): pushing off with one skate while gliding on the other, without relying on human motion data, imitation learning, or precomputed trajectories. We train a control policy in a massively parallelized simulation environment and successfully deploy it on the Booster T1 humanoid robot~\cite{BoosterTech} equipped with standard commercial inline skates.

The main contributions of this paper are as follows:
\begin{itemize}[noitemsep, topsep=2pt, partopsep=0pt]
    \item The hardware modification and integration of completely passive consumer inline skates onto a bipedal humanoid platform using custom-designed \ac{sla} mounts.
    \item A reinforcement learning pipeline
    that circumvents simulation contact artifacts by using distinct geometric primitives (spherical and ellipsoidal) for wheel modeling during training and validation.
    \item A kinematic wheel-rolling reward formulation that incentivizes gliding, synthesizing a feasible skating policy entirely without the need for pre-programmed foot trajectories or motion priors.
    \item The successful sim-to-real deployment of the learned policy, demonstrating the ability to execute fluid skating motions combining propulsion and turning across multiple distinct floor surfaces.
\end{itemize}


\section{Related work} \label{related}

\subsection{Hardware Configurations for Robotic Skating}
Previous attempts to leverage the energy efficiency of skating with passive wheels in robotics have been largely limited to quadrupedal platforms that utilize specialized hardware modifications, such as static feet that can transform into passive wheels~\cite{Endo01052012}, or high-friction claws to push off from~\cite{Bjelonic2018Skating}, which demonstrated promising reductions in \ac{cot} of up to 80\%. Attempts to translate these efficiency gains to bipedal inline skating explored specific mechanical designs~\cite{DesignGaitCastor}, while other skating bipedal robots have circumvented the difficulty of passive propulsion entirely by using a combination of partially actuated and unactuated wheels~\cite{Disney2023StorytellingThroughCharactersSXSW}. Concurrent with our work, Gu et al.~\cite{Gu2026SKATER} recently introduced SKATER, a custom 25-DoF humanoid robot with integrated passive wheels designed specifically for inline skating. In contrast to these custom-built hardware solutions, our approach tackles the integration of off-the-shelf consumer inline skates onto a general-purpose humanoid, requiring the control policy to manage the full physical complexity of unmodified skates.

\subsection{Control Strategies for Passive Wheels}
Managing the inherent instability of single-line passive wheels poses a significant challenge. Unlike actuated wheeled-legged systems that propel themselves through mostly stable driving configurations~\cite{Bjelonic_2019}, bipedal inline skating requires precise control over not only the position of the skates, but also the pitch, roll, and yaw angles of the wheels to properly engage the edges for propulsion and braking~\cite{skate_physics}, demanding 6 \ac{dof} for the skate end effector. Furthermore, the system faces significant partial observability; without encoders on the passive wheels, the controller must infer wheel speed and slippage through secondary sensors, complicating state estimation. 

To mitigate this, early control strategies relied on classical frameworks, such as Zero Moment Point (ZMP) \cite{zmp35}. However, these methods often require explicit phase modeling and depend heavily on forward ground reaction forces, limiting their applicability to the non-holonomic constraints of single-line skates. Consequently, these approaches enforced conservative constraints, such as keeping all passive wheels in continuous contact with the ground and avoiding a swing leg to maximize stability margins~\cite{Kimura2020}. The SKATER robot~\cite{Gu2026SKATER} similarly utilizes a deep \ac{rl} framework where specific reward formulations, such as inter-foot distance constraints and limb symmetry, encourage the policy to converge on a continuous-contact swizzle gait in order to ensure performance and stability. In contrast, our approach utilizes friction generated by wheel edges for true dynamic propulsion. Our wheel modeling validation technique successfully ensures that our policy does not exploit simulation artifacts caused by wheel edge contacts with the ground, allowing the \ac{rl} policy to learn a skating strategy without relying on continuous contact restrictions. Previous skating robots that have utilized strategies involving pushing off wheel edges used precomputed open loop kinematic control~\cite{zivSkate}, but these methods lack the dynamic robustness to reject disturbances, and must be adapted every time for different desired trajectories.

\subsection{Reinforcement Learning for Dynamic Locomotion}
Driven by advances in simulation environments, \ac{rl} has enabled humanoid robots to achieve robust performance in complex whole-body movement tasks, such as martial arts~\cite{xie2025kungfubot} and table tennis~\cite{su2025hitterhumanoidtabletennis}. Closely related to our domain, skateboarding humanoid robots have also shown promising results using \ac{rl}. Previous control strategies for skateboarding that utilized precomputed open-loop motions~\cite{TaskasugiSkate} suffered from increasing complexity when attempting to incorporate turning while pushing and recovering from disturbances. However, training \ac{rl} policies in massively parallelizable environments have proven feasible for skateboarding both in simulation and in real-world deployment~\cite{HUSKY}.

We extend these \ac{rl} methods to the 6 \ac{dof} control required for bipedal inline skating. We furthermore relax the continuous-contact constraints used in prior work, allowing our \ac{rl} policy to learn dynamic balancing and edge control necessary for more complex maneuvers. This enables learned behaviors such as human-like stroking, turning while accelerating, and rapid deceleration.


\section{MECHANICAL DESIGN} \label{hardware}

The hardware platform utilized is the Booster T1, a humanoid robot with 23 DoF. The leg kinematics are well suited for skating, with 6-DoF legs: 3-DoF hips, 1-DoF knees, and 2-DoF ankles, allowing precise control over both skate position and wheel edge angles. Our control policy ignores the two degrees of freedom of the head actuators, leaving them in their neutral position to reduce unnecessary complexity. However, our policy still controls the Booster T1's arms for stability and balance recovery, similar to human skating behavior.

\begin{figure}[tbp]
    \centering
    \includegraphics[width=0.75\linewidth]{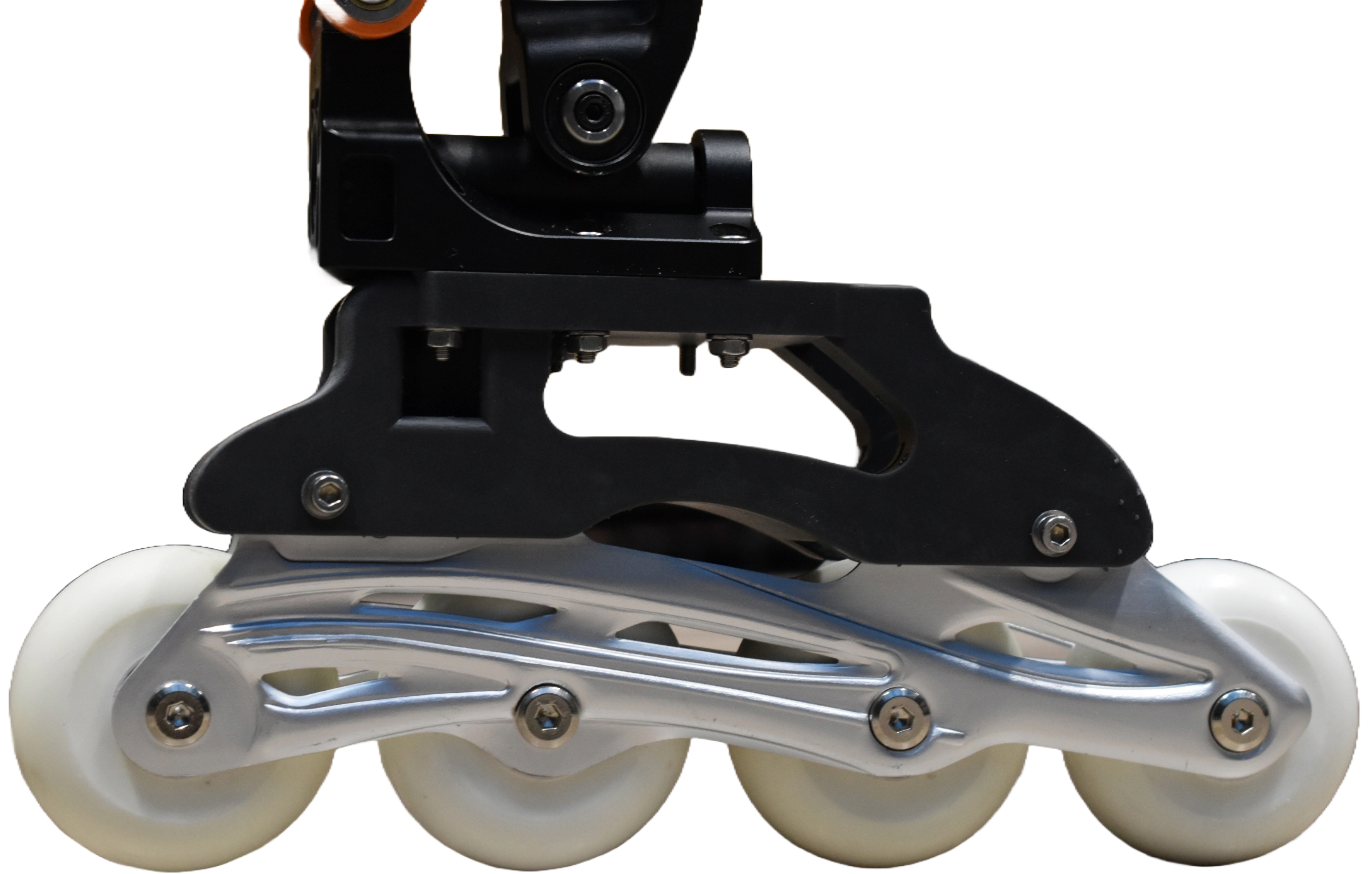}
    \includegraphics[width=0.23\linewidth]{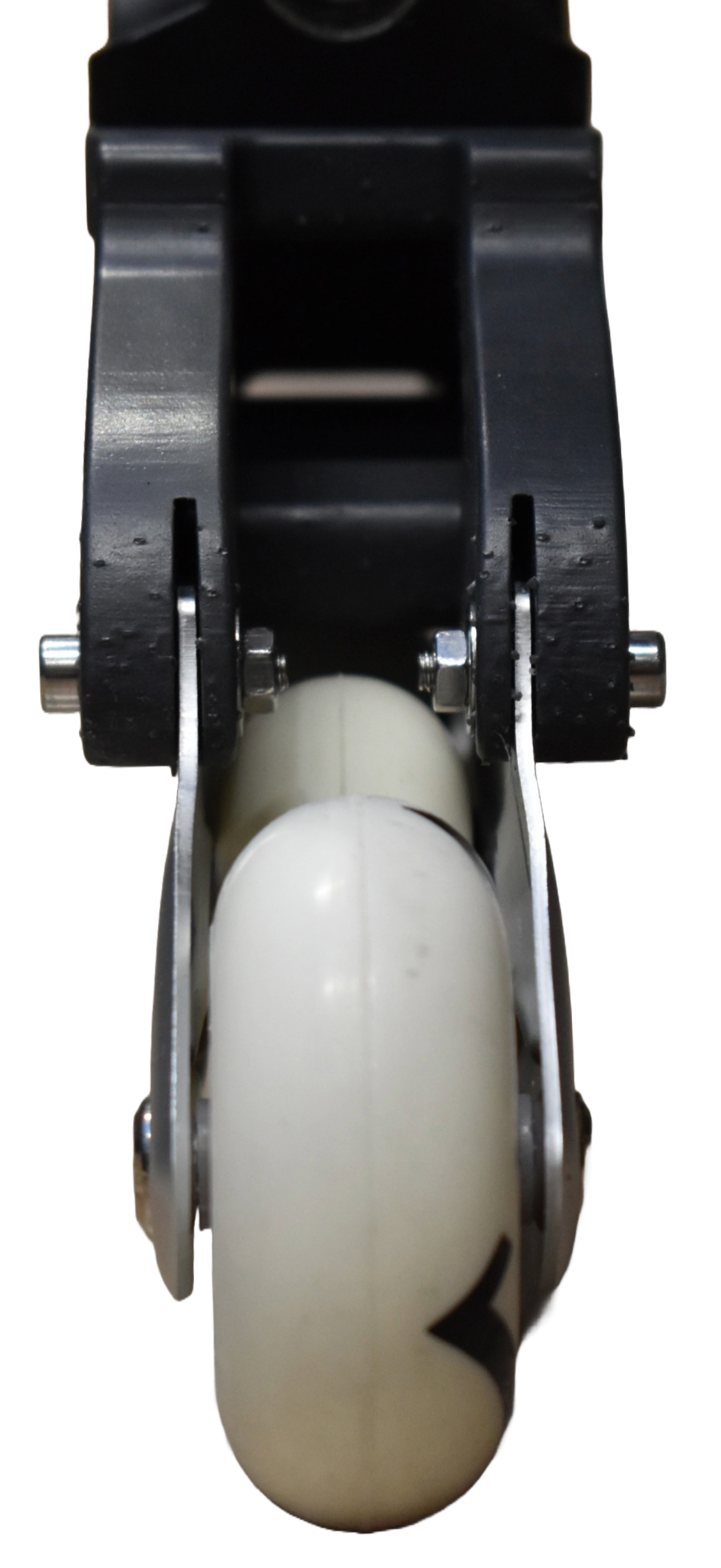}
    \caption{SLA mounts and inline skate bracket and wheels. The mounts are optimized for weight efficiency, with arch designs integrated to resist high speed impacts with the ground with minimal deflection, while maintaining a low material volume.}
    \label{fig:mounts}
\end{figure}

We modify the Booster T1 robot by removing the standard feet and replacing them with a pair of consumer children's inline skates that mimic the proportions used by human inline skaters. The boot of the skates is detached from the metal frame and the wheels, which are slotted directly onto a custom SLA mount that connects to the robot's heel joint (see Fig.~\ref{fig:mounts}). We utilize a special resin, Formlabs' Tough 1500, for its elasticity and ability to withstand the frequent high impact shocks that occur during skating. The skate bearings are substituted for bearings of higher quality (Rollerblade Bearings ILQ-9 Twincam Pro) to minimize friction loss, and we replace the standard wheels with those composed of a less compressible wheel material (Hydrogen Street 60/92A Roller Blade Wheels) to better align with the material properties used in our rigid-body simulation environment.

We utilize the robot's internal sensor package for observations, specifically using proprioceptive feedback from high-resolution dual-encoders on every joint to measure positions and velocities. An \ac{imu} integrated into the torso provides 3-axis angular velocities and accelerations used to estimate the base orientation relative to gravity.

\section{METHOD} \label{training_pipeline}

We train our control policy on a modified version of \emph{Booster Gym}~\cite{boostergym}. We employ an asymmetric actor-critic architecture to train a robust policy capable of zero-shot sim-to-real transfer~\cite{nahrendra2023dreamwaqlearningrobustquadrupedal}, separating the information available during simulated training from what is deployable on the hardware.

\subsection{Problem Formulation}

We model our environment as an infinite-horizon \ac{pomdp}, defined by the tuple $\mathcal{M} = (\mathcal{S}, \mathcal{O}, \mathcal{A}, p, r, \gamma)$. 

We define the continuous full state, partial observation, and action spaces as $\mathbf{s} \in \mathcal{S}$, $\mathbf{o} \in \mathcal{O}$, and $\mathbf{a} \in \mathcal{A}$, respectively. Each episode begins by drawing an initial state from the distribution $d_0(\mathbf{s_0})$. The system then evolves according to the transition dynamics $p(\mathbf{s_{t+1}}|\mathbf{s_t}, \mathbf{a_t})$, yielding a scalar reward defined by $r: \mathcal{S} \times \mathcal{A} \rightarrow \mathbb{R}$ at each timestep. The discount factor is defined by $\gamma \in [0, 1)$. 

We define the agent's goal as learning a policy $\pi(\mathbf{a_t} | \mathbf{o_t})$ that maximizes the expected return:
\begin{equation}
    J(\pi) = \mathbb{E}_{\tau \sim p_\pi} \left[ \sum_{t=0}^{\infty} \gamma^t r(\mathbf{s_t}, \mathbf{a_t}) \right],
\end{equation}
where $\tau = (\mathbf{s_0}, \mathbf{a_0}, \mathbf{s_1}, \dots)$ represents a trajectory of the agent sampled from the POMDP $\mathcal{M}$ under the policy $\pi$. In the context of \ac{ppo}~\cite{schulman2017proximal}, the objective is to find the optimal policy parameters $\theta^*$ such that:
\begin{equation}
    \theta^* = \arg\max_{\theta} J(\pi_\theta)
\end{equation}

\subsection{Observation and Action Spaces}
 Our actor network evaluates the policy using a 74-dimensional observation vector $\mathbf{o_{t}}\in\mathbb{R}^{74}$. This vector is strictly composed of proprioceptive hardware-available sensor measurements alongside internal state variables. We use the notation $\tilde{x}$ to designate a noisy measurement of a true physical quantity $x$. The components of the actor's observation vector are detailed in Table~\ref{tab:observation}.

\begin{table}[htbp]
\caption{Actor Observation Space ($\mathbf{o_{t}}$)}
\label{tab:observation}
\centering
\begin{tabular}{llc}
\hline
\textbf{Component} & \textbf{Symbol} & \textbf{Dimension} \\
\hline
Projected gravity (noisy) & $\tilde{g}_{\text{proj}}$ & 3 \\
Base angular velocity (noisy) & $\tilde{\omega}$ & 3 \\
Velocity commands & $u$ & 3 \\
Gait phase clock & $\Phi$ & 2 \\
Joint positions (noisy) & $\tilde{q}$ & 21 \\
Joint velocities (noisy) & $\dot{\tilde{q}}$ & 21 \\
Previous actions & $a_{t-1}$ & 21 \\
\hline
\textbf{Total Dimension} & & \textbf{74} \\
\hline
\end{tabular}
\end{table}

During training, the critic network estimates the value function $V_{\phi}(\mathbf{s_t})$ using an 88-dimensional state vector $\mathbf{s_{t}}\in\mathbb{R}^{88}$. Crucially, instead of receiving the noise-corrupted observations, the critic's state vector contains the noiseless, ground-truth versions of the actor's observation data, appended with privileged variables that are inaccessible to the physical robot's onboard sensors. The composition of the critic's state vector is summarized in Table~\ref{tab:state}.

\begin{table}[htbp]
\caption{Critic State Space ($\mathbf{s_{t}}$)}
\label{tab:state}
\centering
\begin{tabular}{llc}
\hline
\textbf{Component} & \textbf{Symbol} & \textbf{Dimension} \\
\hline
Projected gravity & $g_{\text{proj}}$ & 3 \\
Base angular velocity & $\omega$ & 3 \\
Velocity commands & $u$ & 3 \\
Gait phase clock & $\Phi$ & 2 \\
Joint positions & $q$ & 21 \\
Joint velocities & $\dot{q}$ & 21 \\
Previous actions & $a_{t-1}$ & 21 \\
Base mass and CoM offset & $m_{\text{base}}, c_{\text{base}}$ & 4 \\
Base linear velocity & $v_{\text{base}}$ & 3 \\
Base height above terrain & ${h}_{\text{base}}$ & 1 \\   
External forces and torques & $F_{\text{ext}}, \tau_{\text{ext}}$ & 6 \\
\hline
\textbf{Total Dimension} & & \textbf{88} \\
\hline
\end{tabular}
\end{table}

The actor outputs an action $\mathbf{a_{t}}\in\mathbb{R}^{21}$ representing the target joint positions. These targets are passed to a low-level \ac{pd} controller operating at a higher frequency to generate the final motor torques~\cite{boostergym}.

\subsection{Contact Modeling for Passive Wheels}
A fundamental challenge in simulating completely passive inline skates is the physical artifacts generated by physics engines. High-resolution meshes suffer from the "sticky contact patch" phenomenon, where multi-point collision resolution artificially dampens rolling motion. Conversely, utilizing standard cylindrical geometric primitives results in edge clipping when the skate tilts to engage its edges. Isaac Gym's PhysX physics engine resolves this clipping by applying artificial restitution impulses, which the RL agent rapidly learns to exploit for cost-free propulsion. Although capsule primitives are commonly used to mitigate sharp-edge clipping, they require a radius and an internal half-height, both of which must be greater than or equal to zero. A capsule with a half-height of zero results in a sphere, while any non-zero half-height introduces an extended linear contact patch along the cylinder that fails to accurately represent the thin profile of an inline skate wheel.

\begin{figure}[tbp]
    \centering
    \includegraphics[width=0.75\linewidth]{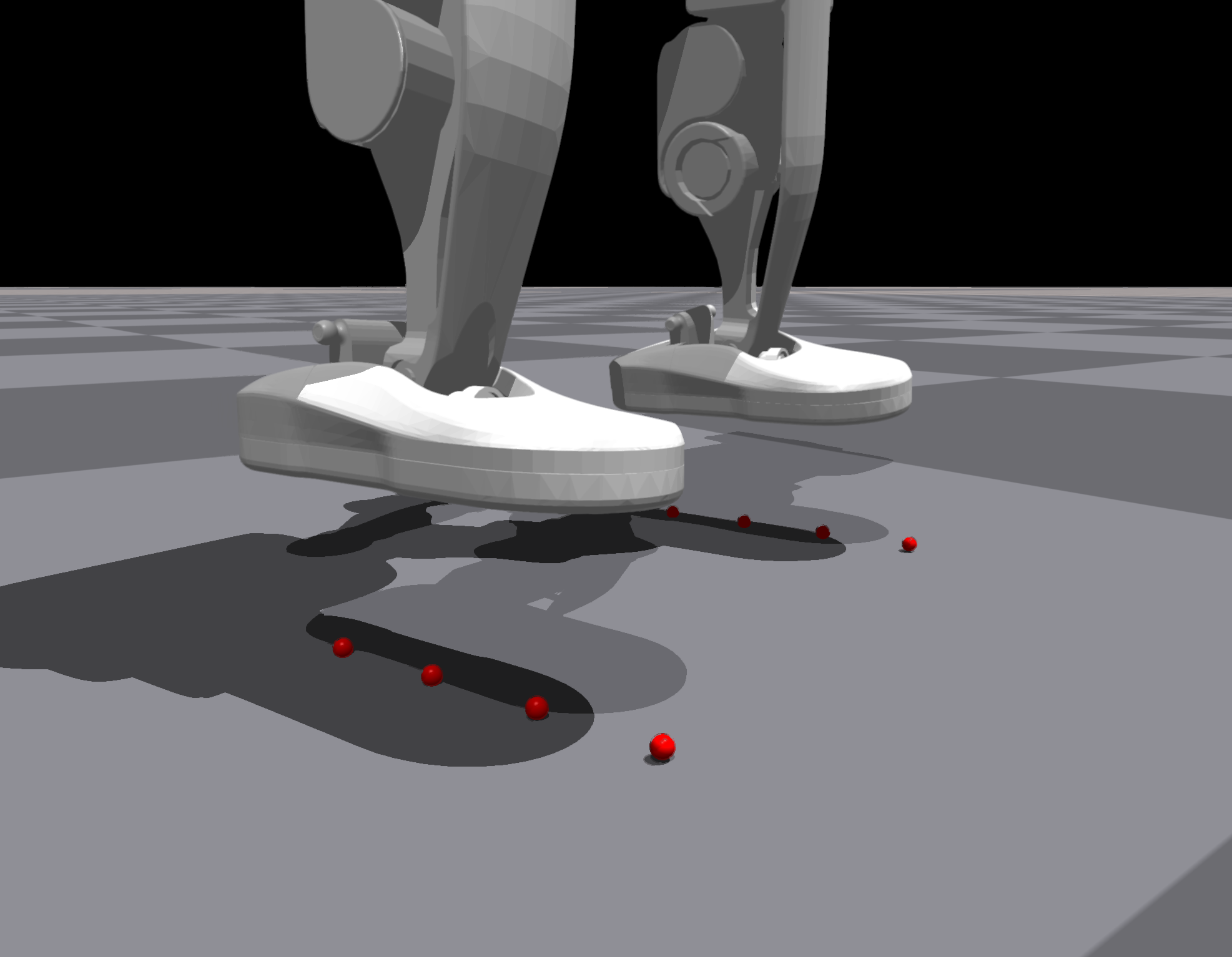}
    \includegraphics[width=0.75\linewidth]{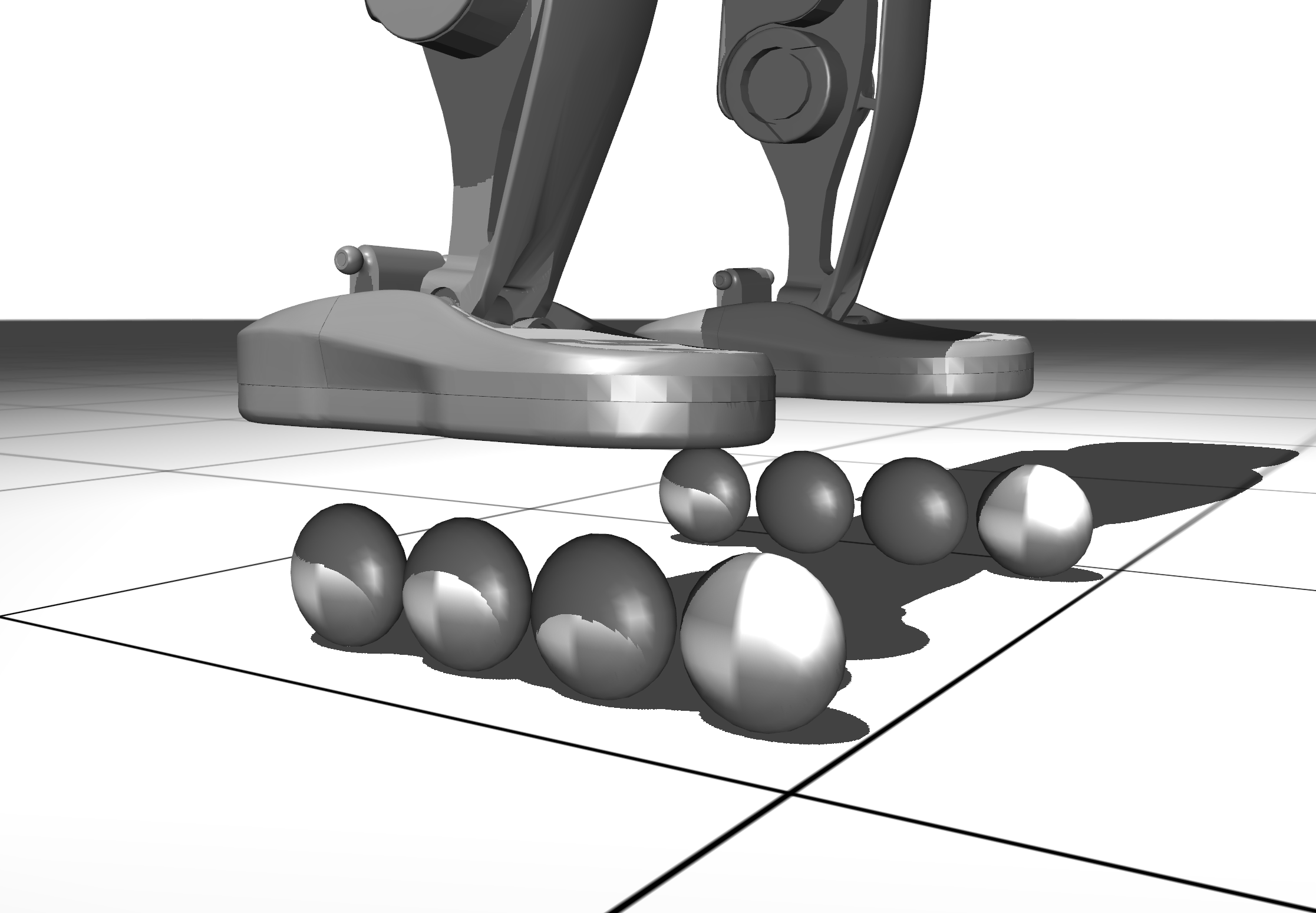}
    \caption{Spherical primitives representing wheel contact points are modeled in Isaac Gym (top). Ellipsoidal wheels are modeled in MuJoCo (bottom).}
    \label{fig:wheels}
\end{figure}

To eliminate these artifacts and ensure dynamic realism, we decouple the wheel geometries used during training and validation. In the Isaac Gym~\cite{makoviychuk2021isaac} training environment, we fit spherical primitives to match the bottom curvature of the physical wheels, ensuring a stable single-point contact patch. To validate the policy and prevent overfitting to a specific geometric abstraction, we employ ellipsoidal primitives during sim-to-sim testing in MuJoCo~\cite{todorov2012mujoco}, as shown in Fig.~\ref{fig:wheels}. A policy demonstrating robustness across both geometric approximations is assumed to generalize well to the true physical geometry of the wheel. Our training pipeline is built within Isaac Gym to maintain direct structural compatibility with the foundational \emph{Booster Gym} environment \cite{boostergym}. While newer simulation frameworks such as Isaac Lab~\cite{nvidia2025isaaclabgpuacceleratedsimulation} utilize updated physics solvers, the core challenges of modeling rigid-body primitive contacts for passive, narrow-profile wheels remain a persistent challenge.

\subsection{Reward Formulation}
Training attempts using standard bipedal reward functions result in a stepping or shuffling behavior, as the agent avoids the instability of rolling. To incentivize authentic skating, we additionally formulate a kinematic rolling reward. 

First, we calculate the average linear velocity of each foot $v_{*}$ by mapping the angular velocities of its wheels $\omega_{n}^{*}$ to linear motion. We isolate this calculation strictly to the wheels currently in contact with the ground:
\begin{equation}
    v_{*}=r\frac{\sum_{n=1}^{N}\omega_{n}^{*}c_{n}^{*}}{\sum_{n=1}^{N}c_{n}^{*}},
\end{equation}
where $r$ is the wheel radius, $N$ is the number of wheels per skate, and $c_{n}^{*}$ is a binary contact indicator ($1$ if the wheel is touching the ground, $0$ otherwise). 

The effective linear wheel velocity of the agent $v_{\text{wheel}}$ is then determined by taking the maximum absolute average velocity between the left and right feet:
\begin{equation}
    v_{\text{wheel}}=\max(|v_{\text{left}}|,|v_{\text{right}}|).
\end{equation}

Finally, the rolling reward utilizes a Gaussian kernel to penalize the squared error between the base's forward linear velocity $v_{x}$ and $v_{\text{wheel}}$:
\begin{equation}
    r_{\text{wheel}}=\exp\left(-\frac{(v_{x}-v_{\text{wheel}})^{2}}{\sigma}\right),
\end{equation}
where $\sigma$ controls the strictness of the velocity matching. In addition to the rolling reward, we apply regularization terms that penalize excessive joint torques, joint velocities, joint accelerations, and base orientation errors to enforce smooth, energy-efficient motions and to protect the physical hardware.

\subsection{Success-Based Command Curriculum}
The inherent instability of passive wheels makes initial exploration highly prone to terminal falls, leading to sparse rewards. We structure the learning process using a success-based command curriculum~\cite{bengio2009curriculum, AutomaticGoal}, similar to~\cite{margolis2024rapid}. The velocity command space is discretized into a 2D grid across linear velocity $v_{x}^{\text{cmd}}$ and angular velocity $\omega_{z}^{\text{cmd}}$, while keeping lateral velocity $v_{y}^{\text{cmd}}=0$ to reflect human inline skating kinematics. Let $B_{i,j}$ represent a specific command bin within this grid, where the index $i$ corresponds to the forward linear velocity bin ($v_x^{\text{cmd}}$) and $j$ corresponds to the yaw angular velocity bin ($\omega_z^{\text{cmd}}$). To ensure continuous coverage of the command space, the final continuous velocity commands are generated by adding uniform jitter to the center of the sampled bin. The agent begins training with commands strictly localized to zero velocity $P(B_{0,0})=1$, to master basic balancing.

At the end of each episode, we evaluate the average velocity tracking performance $\bar{e}_{v}$:
\begin{equation}
    \bar{e}_{v} = \frac{1}{T} \sum_{t=1}^{T} \exp\left(-|v_{x,t} - v_{x}^{\text{cmd}}|\right)
\end{equation}
where $T$ is the total number of steps in the episode and $v_{x,t}$ is the base forward velocity at step $t$. A binary success metric $S \in \{0, 1\}$ is then determined by comparing this performance and the duration of the episode with the thresholds $\tau_{\text{track}}$ and $T_{\text{min}}$:
\begin{equation}
    S = \begin{cases} 
    1, & \text{if } (\bar{e}_{v} > \tau_{\text{track}}) \wedge (T \ge T_{\text{min}}) \\ 
    0, & \text{otherwise} 
    \end{cases}
\end{equation}

When the agent consistently achieves $S=1$ for a sampled bin $B_{i,j}$, the curriculum dynamically updates the probability distribution, unlocking adjacent higher-velocity bins $B_{i+1, j}$ and $B_{i, j \pm 1}$ for sampling. This dynamic boundary expansion allows the policy to incrementally synthesize complex skating behaviors, transitioning from standing to gliding, and eventually to high-speed dynamic turning, without collapsing the value function.

\section{RESULTS} \label{results}

Our custom gym environment and skate modeling methods are validated through (i) an ablation study evaluating key training components via sim-to-sim validation, (ii) a \ac{cot} analysis quantifying locomotion efficiency, and (iii) the successful zero-shot sim-to-real deployment of our policy onto the modified Booster T1 hardware.
Training with a step size of 0.02 seconds, the policy requires approximately 100 million steps to learn to maintain a standing balance and roughly 3 billion steps and 20,000 PPO iterations to master skating across the entire command range, successfully populating all curriculum bins with equal probability. Our training run equates to approximately 1.9 years of simulated experience. 

\subsection{Sim-to-Sim Ablation}

We benchmark our complete pipeline against the following ablated variations (Table \ref{tab:Sim-to-Sim_validation}):
\begin{itemize}
    \item Cylindrical wheels: The policy successfully converges in Isaac Gym but does so by exploiting simulation artifacts; the policy therefore fails to track velocities almost completely in sim-to-sim.
    \item No Curriculum: The policy fails to reach comparable reward levels and cannot track the full velocity command range, ultimately failing to leverage the mechanical efficiency of the wheels.
    \item No Rolling Reward ($r_{wheel}$): The agent learns to step or "shuffle" on the wheels instead of gliding. This results in a high-effort, low-robustness locomotion strategy that poorly translates sim-to-sim.
\end{itemize}

To validate our decoupled wheel geometry approach, we evaluated a baseline policy trained exclusively using standard cylindrical wheel primitives. While this policy achieves high simulated rewards, it does so by exploiting physical solver aberrations; specifically, clipping through the ground plane and harvesting artificial restitution impulses for propulsion. This strategy transfers poorly to sim-to-sim validation in MuJoCo with ellipsoidal wheels, with almost no forward velocity tracking, confirming the necessity of our spherical-to-ellipsoidal training pipeline, as can be seen in Table~\ref{tab:Sim-to-Sim_validation}.

\begin{table}[htbp]
    \caption{Sim-to-Sim Validation: Tracking Performance in MuJoCo Using Ellipsoidal Wheels over 100 Trials \protect\footnotemark}
    \label{tab:Sim-to-Sim_validation}
    \centering
    \renewcommand{\arraystretch}{1.5}
    \setlength{\tabcolsep}{3pt}
    \begin{tabular}{lccc}
        \hline
        \textbf{Method} & \makecell{\textbf{Sim.} \\ \textbf{Time (s)}} & \makecell{\textbf{$v_x$ Tracking} \\ \textbf{Error (m/s)}} & \makecell{\textbf{$\omega_z$ Tracking} \\ \textbf{Error (rad/s)}} \\
        \hline
        Cylindrical Wheels & 60.0 $\pm$ 0.0 & 1.34 $\pm$ 0.33 & 0.56 $\pm$ 0.32 \\
        No Rolling Reward & 55.7 $\pm$ 12.0 & 1.47 $\pm$ 0.28& 0.25 $\pm$ 0.13\\
        No Curriculum & 56.7 $\pm$ 10.1& 0.70 $\pm$ 0.21& 0.13 $\pm$ 0.10\\
        \textbf{Deployed Policy} & \textbf{60.0 $\pm$ 0.0} & \textbf{0.43 $\pm$ 0.26} & \textbf{0.12 $\pm$ 0.08} \\
        \hline
    \end{tabular}
\end{table}

\footnotetext{Commands Sampled From $v_x^{\text{cmd}} \in [1.0,2.0]$ m/s, $\omega_z^{\text{cmd}} \in [-0.5,0.5]$ rad/s for a maximum duration of 60 seconds. Values represent the mean $\pm$ standard deviation.}

Policies trained without our custom rolling reward function also behave poorly, with occasional falls and poor forward velocity tracking. Policies trained without a velocity curriculum perform closer to our deployed policy, but still feature unstable skating techniques that risk potential failure when deployed on the real robot.

\subsection{Cost of Transport Analysis}

\begin{figure}[tbp]
    \centering
    \input{data/CoT_heatmaps} 
    \caption{\ac{cot} values for Booster Gym trained walking policy (top) and our inline skating policy (bottom) at measured body velocities, averaged between Isaac Gym and MuJoCo. Green crosses indicate lower of the two \ac{cot} for that command. We see that while skating is much more energy efficient than walking, the skating policy struggles to turn quickly on the spot, and spans less possible command range than the walking policy. This is in part due to the non holonomic constraints of skating, and in part due to our policy relying on a dominant leg to propel itself forward, hence the asymmetry in \ac{cot} along the yaw velocity axis.}
    \label{fig:CoT}
\end{figure}
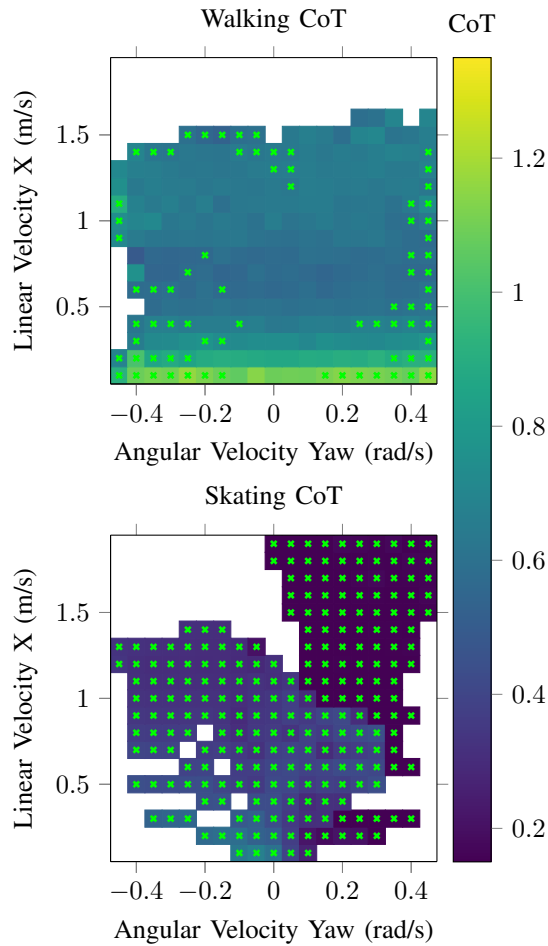

To quantify the efficiency of the learned skating gait, we evaluate the \ac{cot} across various commanded velocities and compare them to that of a conventional walking locomotion policy (Fig.~\ref{fig:CoT}). We define the \ac{cot} as:
\begin{equation}
    \text{CoT} = \frac{\sum_{i=1}^{n} |\tau_i \dot{q}_i|}{m g v_x}
\end{equation}
where $\tau_i$ and $\dot{q}_i$ represent the torque and angular velocity of the $i$-th actuated joint, $m$ is the total mass of the robot, $g$ is the gravitational acceleration, and $v_x$ is the forward velocity of the base.

While walking incurs a lower \ac{cot} at low forward velocity commands where $v_x^{\text{cmd}} <0.5$ m/s, we observe that our skating policy matches and exceeds the walking policy at almost all higher forward velocity intervals. When sampling the velocity command $v_x^{\text{cmd}} = 1$ m/s, $\omega_z^{\text{cmd}} = 0.0$ rad/s, the walking policy results in a \ac{cot} of 0.665, while our skating policy obtains a \ac{cot} of 0.326, showcasing a 50\% decrease in \ac{cot}. At even higher $v_x^{\text{cmd}}$, our policy maintains roughly the same \ac{cot}, while the walking policy collapses, as shown in Fig.~\ref{fig:CoT}.

\begin{figure}[tbp]
    \centering
    \subfloat[Composite image of a left turn at speed. The shown frames are sampled at 1s intervals.]{\centering%
    \includegraphics[width=0.48\textwidth]{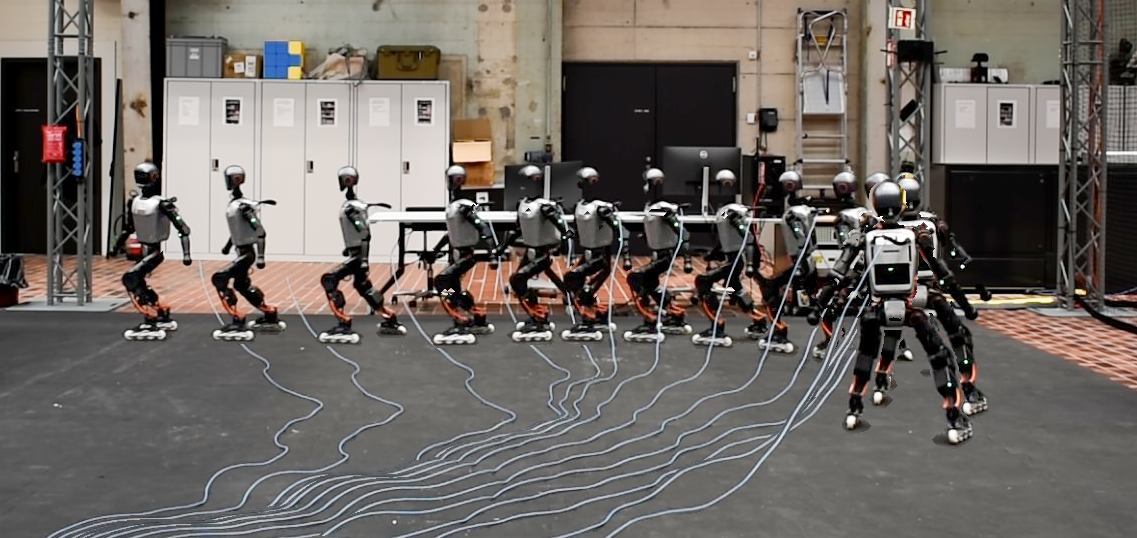}%
    }%
    \\
    \subfloat[Wooden flooring]{\includegraphics[width=0.15\textwidth]{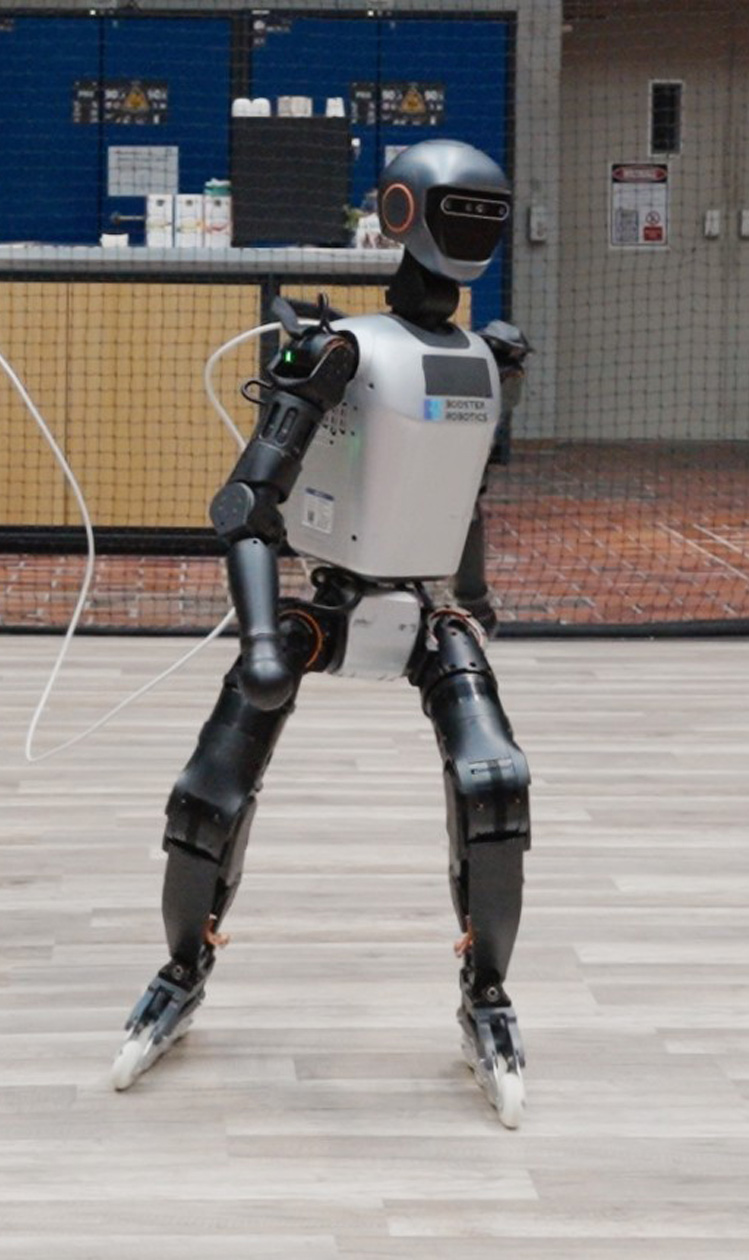}}
    \subfloat[Brick paving]{\includegraphics[width=0.15\textwidth]{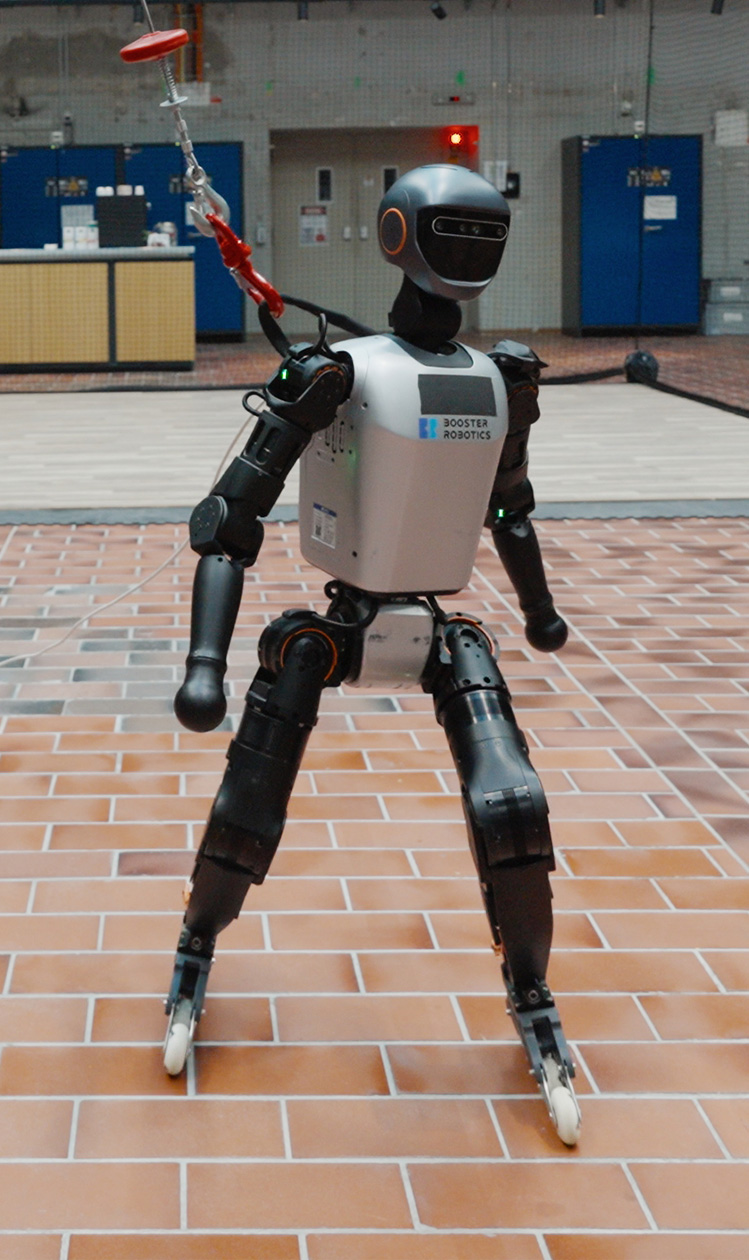}}%
    \subfloat[Padded mat]{\includegraphics[width=0.15\textwidth]{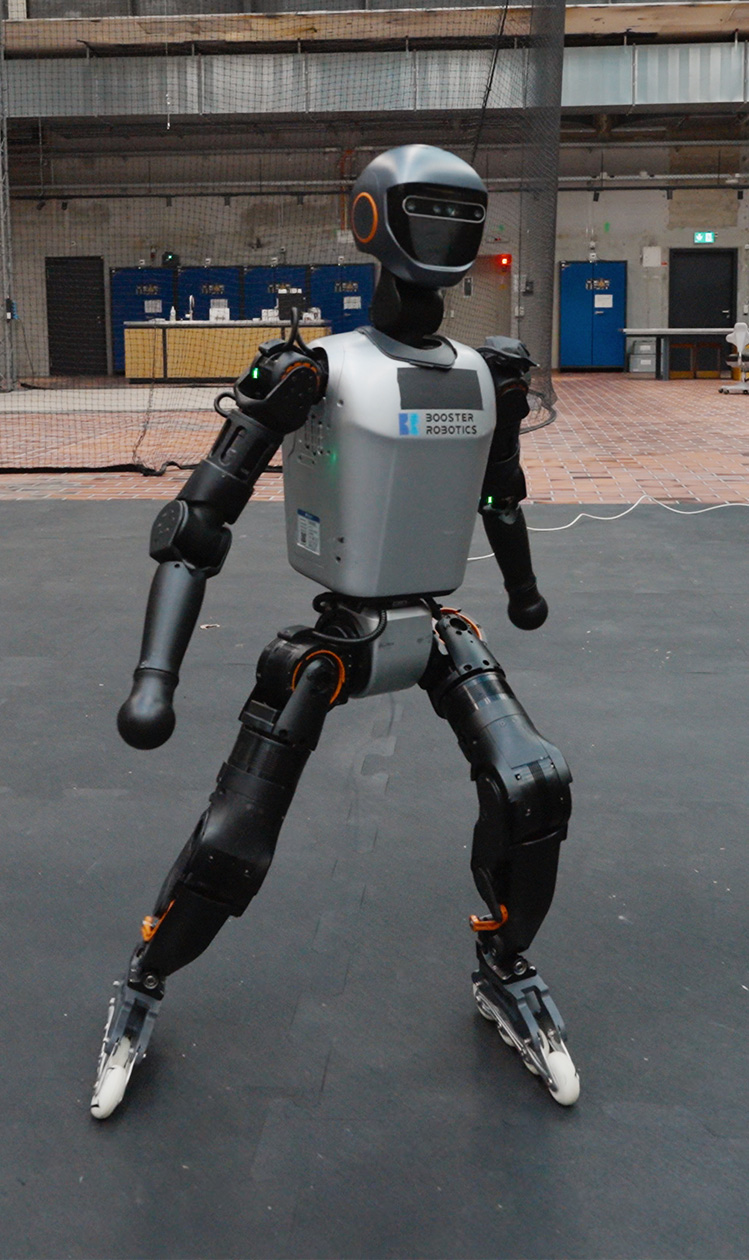}}%
    \caption{Real-world deployment of the learned skating policy turning at speed (a), along with zero-shot robustness across various terrain (b-d), demonstrating our policy's ability to withstand unmodeled variations in surface friction, rolling resistance, and topographical irregularities.}
    \label{fig:deployment}
\end{figure}

\begin{figure*}[tbp]
    \centering
    \includegraphics[width=0.98\linewidth]{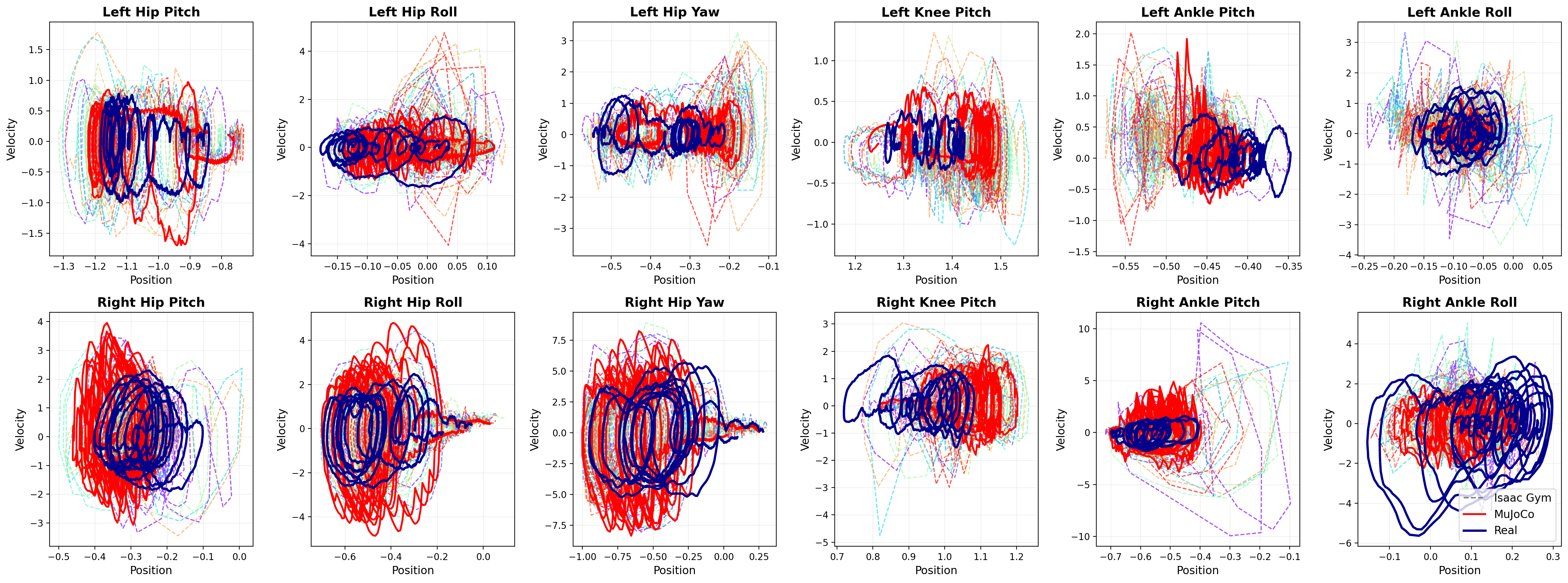}
    \caption{Position-velocity trajectories of leg joints from 0 m/s to 1.8 m/s and back to 0 m/s forward velocity command, with zero yaw commands. Isaac Gym trajectories are aggregated from 10 randomized environments and are shown in dashed lines. Trajectory discrepancies are accentuated by real-world surface unevenness and varying friction coefficients. Because skating relies on continuous ground contact rather than discrete steps, these surface mismatches compound errors and impact our policy significantly more than standard locomotion.}
    \vspace{-2mm}
    \label{fig:deployment_q}
\end{figure*}

\subsection{Hardware Deployment}

\begin{figure}[tbp]
    \centering
    \input{data/CoT_Real_Compare} 
    \caption{\ac{cot} values for the baseline walking policy compared to our inline skating policy during real-world deployment on padded mats. At increased forward velocities, inline skating becomes significantly more energy-efficient and attains higher overall speeds than standard walking.}
    \vspace{-4mm}
    \label{fig:CoT_Real}
\end{figure}
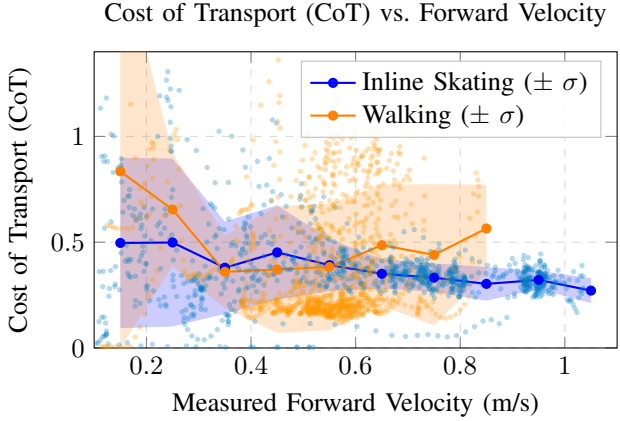

The trained policy successfully transfers to the physical Booster T1 robot, demonstrating dynamic locomotion as shown in Fig.~\ref{fig:deployment_q}. During untethered deployment~\footnote{We utilize an Ethernet cable as an additional safety stop during locomotion, but ensure no resultant tension or external forces are felt by the robot during operation.}, the robot effectively showcases forward acceleration, deceleration, and dynamic turning while in motion (Fig.~\ref{fig:deployment}). Furthermore, the agent is capable of performing zero-radius turns on the spot and exhibits high robustness to active physical perturbations, successfully recovering from continuous external pushes. The agent is also robust to uneven flooring, to which it was never exposed, and to various floor surfaces. Real-world measurements demonstrate that while standard walking exhibits a comparable \ac{cot} at moderate speeds, our trained inline skating policy becomes significantly more efficient past a velocity threshold of approximately 0.6 m/s (Fig.~\ref{fig:CoT_Real}). Furthermore, the skating policy sustains this low \ac{cot} while achieving higher overall forward velocities that walking cannot acheive. Notably, these results were recorded on padded mats, a surface whose cushioning properties are suboptimal for inline skating and entirely out-of-distribution for our simulated training environment.

\section{CONCLUSION} \label{conclusion}

This paper presents a novel reinforcement learning framework that enables a bipedal humanoid robot to achieve dynamic locomotion using passive consumer inline skates. By utilizing distinct geometric wheel primitives across training and validation environments, we circumvent common physics engine artifacts to accurately model passive wheel dynamics, specifically in hard-to-model cases such as pushing off wheel edges. Coupled with a success-based command curriculum and a kinematic rolling reward, the resulting policy successfully bridges the sim-to-real gap in zero-shot deployment. The deployed agent demonstrates robust balancing, dynamic turning, and the ability to reject external disturbances without relying on predefined motion trajectories or human motion data. To the best of our knowledge, this work represents the first successful real-world deployment of a bipedal humanoid robot that executes emergent stroke-and-glide propulsion, paralleling the way humans skate.

However, physical space constraints limit our ability to test the policy's complete velocity command range and accuracy on the hardware. Furthermore, the RL optimization landscape leads the policy to converge on a "dominant leg" propulsion strategy, despite efforts to enforce symmetrical constraints. While this local minimum increases the robot's immediate survivability by minimizing the instability of weight transfers, it deviates from the alternating, high-power strokes characteristic of human skaters, indicating that the resulting gait may still be suboptimal and less energy-efficient.

Future work aims to refine the simulation environment and reward formulation to explicitly incentivize alternating-leg propulsion, while adapting our framework built on \emph{Booster Gym}~\cite{boostergym} to newer simulation frameworks such as Isaac Lab~\cite{nvidia2025isaaclabgpuacceleratedsimulation}. Additionally, expanding this framework to ice skating, which introduces highly complex friction and phase-change dynamics, and developing a control framework capable of autonomously transitioning between standard walking and skating configurations and control policies remain compelling avenues for extending humanoid agility.



\FloatBarrier
\section*{APPENDIX}

This appendix provides the detailed configurations and parameters used in our reinforcement learning framework. The following tables outline the domain randomization ranges applied for robust sim-to-real transfer (Table~\ref{tab:domain_rand}), the PPO hyperparameters and network architectures (Table~\ref{tab:ppo_params}), the core physics engine settings used for the Isaac Gym simulation environment (Table~\ref{tab:sim_params}), and the mathematical formulations and weights for the reward functions (Table~\ref{tab:reward_functions}).
\subsection{Domain Randomization Parameters}
\vspace{-1.5mm}
\begin{table}[htbp]
\vspace{-2mm}
\caption{Domain Randomization Parameters}
\vspace{-2mm}
\label{tab:domain_rand}
\centering
\footnotesize
\renewcommand{\arraystretch}{0.95}
\begin{tabular}{ll}
\toprule
\textbf{Parameter} & \textbf{Range} \\
\midrule
Init. DoF Position (Additive) & $[-0.2, 0.2]$ rad \\
Init. Base Pos (X, Y) (Additive)& $[-1.0, 1.0]$ m \\
Init. Base Lin. Vel. (X, Y) (Additive)& $[0.0, 0.2]$ m/s \\
Kick Linear Velocity (Additive)& $[0.0, 0.1]$ m/s \\
Kick Angular Velocity (Additive)& $[0.0, 0.02]$ rad/s \\
Push Force (Additive)& $[0.0, 5.0]$ N \\
Push Torque (Additive)& $[0.0, 0.5]$ Nm \\
DoF Stiffness (Scaling)& $[0.95, 1.05]$ \\
DoF Damping (Scaling)& $[0.95, 1.05]$ \\
DoF Friction (Additive)& $[0.0, 2.0]$ \\
Ground Friction (Additive)& $[-0.5, 0.5]$ \\
Wheel Friction (Additive)& $[-0.3, 3.0]$ \\
Wheel DoF Friction (Additive)& $[-0.0001, 0.0001]$ \\
Compliance (Additive)& $[0.5, 1.5]$ \\
Restitution (Additive)& $[0.1, 0.9]$ \\
Base CoM (Additive)&$[-0.1, 0.1]$ m \\
Base Mass (Scaling)& $[0.8, 1.2]$ \\
\bottomrule
\end{tabular}
\end{table}

\FloatBarrier
\vspace{2cm}
\subsection{PPO and Simulation Parameters}

\begin{table}[htbp]
    \centering
    \begin{minipage}[t]{0.48\columnwidth}
        \centering
        \caption{PPO Hyperparameters}
        \label{tab:ppo_params}
        \resizebox{\linewidth}{!}{%
        \begin{tabular}{lc}
        \toprule
        \textbf{Parameter} & \textbf{Value} \\
        \midrule
        Learning Rate & $1.0 \times 10^{-5}$ \\
        Discount Factor ($\gamma$) & 0.995 \\
        GAE Parameter ($\lambda$) & 0.95 \\
        Desired KL Divergence & 0.01 \\
        Entropy Coefficient & -0.01 \\
        Horizon Length & 24 \\
        Mini Epochs & 20 \\
        Actor Architecture & [256, 128, 128] \\
        Critic Architecture & [256, 256, 128] \\
        \bottomrule
        \end{tabular}%
        }
    \end{minipage}\hfill
    \begin{minipage}[t]{0.48\columnwidth}
        \centering
        \caption{Simulation Parameters}
        \label{tab:sim_params}
        \resizebox{\linewidth}{!}{%
        \begin{tabular}{lc}
        \toprule
        \textbf{Parameter} & \textbf{Value} \\
        \midrule
        Number of Environments & 8092 \\
        Sim. Time Step ($\Delta t$) & 0.002 s \\
        Substeps & 1 \\
        Physics Engine & PhysX \\
        PhysX Solver & TGS \\
        Position Iterations & 4 \\
        Velocity Iterations & 4 \\
        Contact Offset & 0.003 m \\
        Bounce Threshold Vel. & 2.0 m/s \\
        Max Depenetration Vel. & 0.5 m/s \\
        Max GPU Contact Pairs & $8,388,608$ \\
        Terrain Static Friction & 1.6 \\
        Terrain Dynamic Friction & 1.5 \\
        \bottomrule
        \end{tabular}%
        }
    \end{minipage}
\end{table}

\vspace{-3mm}
\FloatBarrier
\vspace{-3mm}
\subsection{Reward Functions}
\vspace{-3mm}
\begin{table}[htbp]
\vspace{-2mm}
\caption{Reward Function Definition and Weights}
\vspace{-2mm}
\label{tab:reward_functions}
\centering
\footnotesize
\renewcommand{\arraystretch}{0.95}
\resizebox{\columnwidth}{!}{%
\begin{tabular}{llc}
\toprule
\textbf{Reward Term} & \multicolumn{1}{c}{\textbf{Formula}} & \textbf{Weight} \\
\midrule
\multicolumn{3}{l}{\textit{\textbf{Task Rewards}}} \\
Tracking Lin. X & $\exp(-(v_{x}^{\text{cmd}}-v_{x})^{2}/\sigma_{\text{vel}})$ & $5.0$ \\
Tracking Lin. Y & $\exp(-(v_{y}^{\text{cmd}}-v_{y})^{2}/\sigma_{\text{vel}})$ & $0.5$ \\
Tracking Ang. Z & $\exp(-(\omega_{z}^{\text{cmd}}-\omega_{z})^{2}/\sigma_{\text{ang}})$ & $5.0$ \\
Wheel Spin & $\exp(-(v_{x}^{\text{cmd}}-v_{\text{wheel}})^{2}/\sigma_{\text{vel}})\cdot 1_{\text{contact}}$ & $6.0$ \\
Feet Swing & $(1_{\text{swing},L}\cdot 1_{\neg\text{contact},L}) + (1_{\text{swing},R}\cdot 1_{\neg\text{contact},R})$ & $3.0$ \\
Survival & $1.0$ & $0.05$ \\
\addlinespace
\multicolumn{3}{l}{\textit{\textbf{Base Penalties}}} \\
Lin. Vel. Z & $(v_{z})^{2}$ & $-1.0$ \\
Ang. Vel. XY & $||\omega_{x,y}||^{2}$ & $-0.1$ \\
Root Accel. & $||\dot{v}_{\text{base}}||^{2} + ||\dot{\omega}_{\text{base}}||^{2}$ & $-1 \cdot 10^{-4}$ \\
\addlinespace
\multicolumn{3}{l}{\textit{\textbf{Posture Penalties}}} \\
Base Height & $(h_{\text{base}}-h_{\text{target}})^2$ & $-10.0$ \\
Orientation & $||g_{\text{proj},x,y}||^{2}$ & $-1.0$ \\
\addlinespace
\multicolumn{3}{l}{\textit{\textbf{Joint Penalties}}} \\
Feet Roll & $\sum(q_{\text{footroll}}-q_{\text{footroll},\text{home}})^{2}$ & $-0.8$ \\
Arms Home & $\sum(q_{\text{arm}}-q_{\text{arm},\text{home}})^{2}$ & $-0.5$ \\   
DOF Limits & $\sum(1_{q<q_{\text{low}}} + 1_{q>q_{\text{high}}})$ & $-4.0$ \\
\addlinespace
\multicolumn{3}{l}{\textit{\textbf{Control Penalties}}} \\
Action Rate & $\sum(a_{t}-a_{t-1})^{2}$ & $-1.0$ \\
Torques & $\sum(\tau_{i}\cdot 1_{\text{nonwheel}})^{2}$ & $-2 \cdot 10^{-4}$ \\
Torque Limits & $\sum(\max(0, |\tau_{i}|-\tau_{\text{limit},\text{soft}}))$ & $-2.0$ \\
Ankle Vel. & $\sum\dot{q}_{\text{ankle}}^{2}$ & $-2.5 \cdot 10^{-2}$ \\
DOF Vel. & $\sum\dot{q}_{i}^{2}\cdot 1_{\text{nonwheel}}$ & $-2 \cdot 10^{-4}$ \\
DOF Accel. & $\sum(\ddot{q}_{i})^{2}\cdot 1_{\text{nonwheel}}$ & $-2 \cdot 10^{-7}$ \\
Power & $\sum(\max(0, \tau_{i}\cdot \dot{q}_{i})\cdot 1_{\text{nonwheel}})$ & $-2 \cdot 10^{-3}$ \\
Tiredness & $\sum((\tau_{i}/\tau_{\text{limit}})^{2}\cdot 1_{\text{nonwheel}})$ & $-1 \cdot 10^{-2}$ \\
\addlinespace
\multicolumn{3}{l}{\textit{\textbf{Contact Penalties}}} \\
Collision & $\sum(1_{F_{\text{contact}}>1.0})$ & $-1.0$ \\
\bottomrule
\end{tabular}%
}
\end{table}
\vspace{-2mm}
\FloatBarrier

\vspace{-3mm}

\section*{ACKNOWLEDGMENT}
The authors thank Daniel Wagner for developing the experiment space and Per Frivik for his work on surrogate sphere wheel contact modeling.
\vspace{-2mm}
\bibliographystyle{IEEEtran}
\bibliography{bibliography}

\end{document}

%% file: data/CoT_heatmaps.tex
\begin{tikzpicture}
    \begin{groupplot}[
        group style={
            group size=1 by 2,
            vertical sep=2 cm,
        },
        view={0}{90},
        width=0.5\linewidth,
        height=0.5\linewidth,
        scale only axis,
        enlargelimits=false,
        xlabel={Angular Velocity Yaw (rad/s)},
        ylabel={Linear Velocity X (m/s)},
        colormap/viridis,
        point meta min=0.15,
        point meta max=1.35,
        unbounded coords=jump,
        mesh/cols=20,
        mesh/rows=20,
        mesh/ordering=x varies,
        tick align=outside 
    ]

        \nextgroupplot[
            title={Walking CoT},
            colorbar,
            colorbar style={
                height=10.64 cm,
                at={(1.05, 1)},
                anchor=north west,
                title={CoT},
                ylabel style={yshift=0.5cm},
                tick pos=right
            }
        ]
        \addplot3[
            surf,
            shader=flat corner,
            point meta=explicit,
        ] table [
            col sep=comma,
            x=ang_vel_yaw,
            y=lin_vel_x,
            z expr={0},
            meta=COT_mech
        ] {data/walking_cot_new.csv};
        
        \addplot[
            only marks,
            mark=x,
            mark size=1.5pt,
            mark options={draw=green, line width=1pt}, 
        ] table [
            col sep=comma,
            x=ang_vel_yaw,
            y=lin_vel_x,
        ] {data/walking_wins_new.csv};

        \nextgroupplot[
            title={Skating CoT}
        ]
        \addplot3[
            surf,
            shader=flat corner,
            point meta=explicit,
        ] table [
            col sep=comma,
            x=ang_vel_yaw,
            y=lin_vel_x,
            z expr={0},
            meta=COT_mech
        ] {data/skating_cot_new.csv};
  
        \addplot[
            only marks,
            mark=x,
            mark size=1.5pt,
            mark options={draw=green, line width=1pt}, 
        ] table [
            col sep=comma,
            x=ang_vel_yaw,
            y=lin_vel_x,
        ] {data/skating_wins_new.csv};
        
    \end{groupplot}
\end{tikzpicture}

%% file: data/CoT_Real_Compare.tex
\begin{tikzpicture}
\begin{axis}[
    title={Cost of Transport (CoT) vs. Forward Velocity},
    xlabel={Measured Forward Velocity (m/s)},
    ylabel={Cost of Transport (CoT)},
    xmin=0.1, xmax=1.1,
    ymin=0, ymax=1.4,
    width=8.5cm, 
    height=5.5cm,
    grid=both,
    grid style={dashed, gray!30},
    legend pos=north east,
    legend cell align={left},
    legend style={draw=gray!50, fill=white, fill opacity=0.9, text opacity=1}
]

\addplot[only marks, mark=*, mark size=0.8pt, color=cyan!60!blue, opacity=0.3, forget plot] 
    table[x=vx, y=cot, col sep=comma] {data/skate_scatter.csv};

\addplot[only marks, mark=*, mark size=0.8pt, color=orange!80!yellow, opacity=0.3, forget plot] 
    table[x=vx, y=cot, col sep=comma] {data/walk_scatter.csv};

\addplot[name path=skate_lower, draw=none, forget plot] table[x=vx, y=cot_lower, col sep=comma] {data/skate_binned.csv};
\addplot[name path=skate_upper, draw=none, forget plot] table[x=vx, y=cot_upper, col sep=comma] {data/skate_binned.csv};
\addplot[blue, fill opacity=0.2, forget plot] fill between[of=skate_lower and skate_upper];
\addplot[color=blue, mark=*, mark size=1.5pt, thick] 
    table[x=vx, y=cot_mean, col sep=comma] {data/skate_binned.csv};

\addplot[name path=walk_lower, draw=none, forget plot] table[x=vx, y=cot_lower, col sep=comma] {data/walk_binned.csv};
\addplot[name path=walk_upper, draw=none, forget plot] table[x=vx, y=cot_upper, col sep=comma] {data/walk_binned.csv};
\addplot[orange, fill opacity=0.2, forget plot] fill between[of=walk_lower and walk_upper];
\addplot[color=orange, mark=*, mark size=1.5pt, thick] 
    table[x=vx, y=cot_mean, col sep=comma] {data/walk_binned.csv};

\legend{Inline Skating ($\pm$ $\sigma$), Walking ($\pm$ $\sigma$)}
\end{axis}
\end{tikzpicture}
\label{fig:cot_comparison}